\documentclass{article}

\usepackage{microtype}
\usepackage{graphicx}
\usepackage{subfigure}
\usepackage{booktabs}

\usepackage{hyperref}

\usepackage{microtype}
\usepackage{amsfonts}
\usepackage{graphicx}
\usepackage {amsmath}
\usepackage{booktabs}

\usepackage[accepted]{icml2023}

\usepackage{amsmath}
\usepackage{amssymb}
\usepackage{mathtools}
\usepackage{amsthm}

\usepackage{hyperref}
\usepackage{colortbl}
\usepackage{times,graphicx,amsmath,amssymb,booktabs,tabulary,multirow,overpic,xcolor}
\usepackage[capitalize,noabbrev]{cleveref}

\usepackage[textsize=tiny]{todonotes}

\icmltitlerunning{Random Weights Networks Work  as Loss Prior Constraint}

\begin{document}
	\twocolumn[
	\icmltitle{Random Weights Networks \\ Work  as Loss Prior Constraint for Image Restoration}
	
	
	
	\icmlsetsymbol{equal}{*}

	\begin{icmlauthorlist}
		\icmlauthor{Man Zhou}{equal,1}
		\icmlauthor{Naishan Zheng}{equal,2}
		\icmlauthor{Jie Huang}{2}
		\icmlauthor{Xiangyu Rui}{3}
		\icmlauthor{Chunle Guo}{4}
		\icmlauthor{Deyu Meng}{3}
		\icmlauthor{Chongyi Li}{1}
		\icmlauthor{Jinwei Gu}{5}
	\end{icmlauthorlist}
	
	\icmlaffiliation{1}{S-Lab, Nanyang Technological University, Singapore} 
	\icmlaffiliation{2}{University of Science and Technology of China, China} 
	\icmlaffiliation{3}{Xi'an Jiaotong University, China}
	\icmlaffiliation{4}{Nankai University, China} 
	\icmlaffiliation{5}{SenseTime Research, USA} 
	
	\icmlcorrespondingauthor{Jinwei Gu}
	
	\icmlkeywords{Machine Learning, ICML}
	
	\vskip 0.3in
	]
	
	
	
	\printAffiliationsAndNotice{\icmlEqualContribution} 
	
	\begin{abstract}
		
		In this paper, orthogonal to the existing data and model studies, we instead resort our efforts to investigate the potential of loss function in a new perspective and present our belief \textbf{``Random Weights Networks can Be Acted as Loss Prior Constraint for Image Restoration''}. Inspired by Functional theory, we provide several alternative solutions to implement our belief in the strict mathematical manifolds including Taylor's Unfolding Network, Invertible Neural Network, Central Difference Convolution and Zero-order Filtering  as ``random weights network prototype'' with respect of the following four levels: 1) the different random weights strategies; 2) the different network architectures, \emph{eg,} pure convolution layer or transformer; 3) the different network architecture depths; 4) the different numbers of random weights network combination.  Furthermore, to enlarge the capability of the randomly initialized manifolds, we devise the manner of random weights in  the following two variants: 1) the weights are randomly initialized only once during the whole training procedure; 2) the weights are randomly initialized at each training iteration epoch. Our propose belief can be directly inserted into existing networks without any training
		and testing computational cost. Extensive experiments across multiple image restoration tasks, including image de-noising, low-light image enhancement, guided image super-resolution demonstrate the consistent performance gains obtained by introducing our belief. To emphasize, our main focus is to spark the realms of loss function and save their current neglected status. Code will be publicly available. 
	\end{abstract}

	\section{Introduction}
	\label{submission}
	
	As well recognized, image restoration has long been an important task in
	computer vision field, which aims to recover a latent clear image from a given degraded observation. However, it is highly ill-posed, challenging and remains to be solved as there exists infinite feasible results for single degraded image \cite{9638340,NEURIPS2021_9e3cfc48}. Therefore, image restoration  has received the great interests from computer vision community. The representative image restoration tasks include image de-noising \cite{Zheng_2021_CVPR,9454311,8306131,7839189}, low-light image enhancement \cite{9420270,9879599,9340611}  and guided image super-resolution \cite{9165231,9879770,9662053}.    
	
	Much research efforts have been devoted to solve the single image restoration problem, which can be categorized into two groups: traditional optimization methods \cite{Gong_2016_CVPR,7473901,shi2015lrtv} and deep learning based methods \cite{NEURIPS2021_9e3cfc48,9879770,9662053}. In terms of traditional image restoration methods, researchers have formulated image restoration as an optimization problem and developed various natural image priors to regularize the solution space of the latent clear image, \emph{eg}, low-rank prior \cite{7473901,shi2015lrtv}, dark channel prior \cite{pan2016blind,pan2017deblurring,8100221}, graph-based prior \cite{li2019single,8488519}, total variation regularization \cite{661187,du2018single,2009Variational} and sparse image priors~\cite{li2021comprehensive,yang2020single,s1,s2}. However, these priors need to be carefully designed and these traditional methods involve the iteration optimization, thus consuming the huge computational resources and further hindering their usage. In a word, the common sense is to explore the potential image prior to relieve the optimization difficulty of the ill-posed image restoration.

	On the line of deep learning-based methods, deep neural networks (CNNs) have received widespread attention and achieved promising improvement in image restoration tasks over traditional methods \cite{Xu-nips,zhang2018dynamic,gao2019dynamic,deblurganv2,tao2018scale,deblurgan,Zhang_2018_CVPR,zhang2020residual,fan2020neural,ren2021deblurring}.
	The pioneering image restoration work based on deep neural network belongs to image de-noising orientated DNCNN \cite{7839189}. It stands on the numerous training data and incorporates \textbf{deep learning modeling paradigm} of the powerful feed-forward mapping ability and backward gradient optimization strategies \cite{DBLP:journals/corr/Ruder16}, thus obtaining the remarkable achievement. Since then, explosive deep network architectures have been constructed and struck the image restoration field with the significant performance gain.

	\begin{figure*}[t]
		\setlength{\abovecaptionskip}{-0.3cm}
		\setlength{\belowcaptionskip}{-0.3cm}
		\begin{center}
			\begin{tabular}[t]{c} 
				\includegraphics[width=\textwidth]{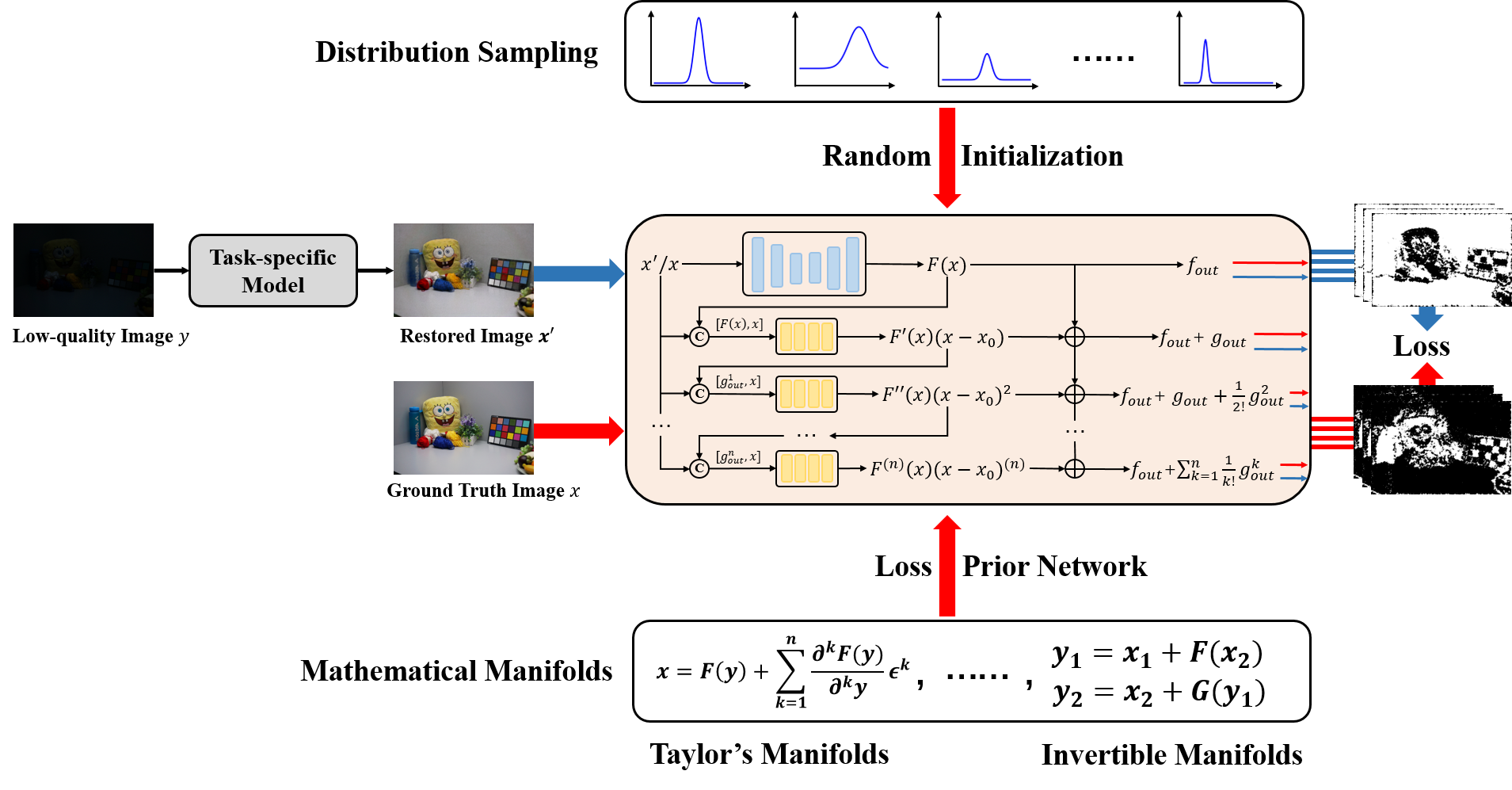}
			\end{tabular}
		\end{center}
		\caption{\small The flowchart of our belief \textbf{``Random Weights Networks can Be Acted as Loss Prior Constraint for Image Restoration''}. In detail, the output of task-specific model and the ground truth are fed into the loss network for loss calculating where the loss network is with strict mathematical manifold and its weights are randomly initialized.}
		\label{fig:pipeline}
	\end{figure*}

	As well recognized, the aforementioned \textbf{deep learning modeling paradigm} involves with three key components: \textbf{data}, \textbf{model} and \textbf{loss function}. Currently, much more efforts have been devoted to the first two ones while seldom studies focus on the remaining loss function. 
	
	\textbf{Related Work.} In terms of the prior loss function studies, the representative one \cite{9157622}  customized a dual regression scheme for image super-resolution task in the form of loss regularization term by introducing an additional constraint on low-resolution data to reduce the space of the possible image super-resolution solutions. Similar to the loss regularization function, CycleGAN framework \cite{9420312,9317750,9154550} exploited two sets of parallel generative  adversarial  networks to formulate the image restoration function and the image degradation mechanism respectively where the corresponding cycle mechanism is modeled in the loss function by the form of regularization term.  In addition,     the work \cite{wang2023gpsr} explored the Range-Null space decomposition to enable the relationship between realness and data consistency and the consistency constraint is transferred into loss function. Despite the remarkable progress, the first two ones have been trained in a delicate manner while the remaining one only works on the particular forms of known degradation matrix. These issues leave a zoom for further study over loss function. 
	
	Different from the existing studies, we instead resort our efforts to investigate the potential of loss function in a new perspective where random weights network is treated as loss function without need of training and hope that this study will spark the realms of loss function community.

	In this paper, we first investigate the importance of loss function and present our belief \textbf{``Random Weights Networks can Be Acted as Loss Prior Constraint for Image Restoration''}. There exists a question ``whether any network architecture with random weights can be served ?". The answer is ``NO''!

	To implement our belief, we provide several simple and alternative solutions of random weights networks in the strict mathematical property and believe the future of much more feasible solutions. Specifically, we stand on the Functional theory and develop the following functional manifolds including Taylor's Unfolding Network, Invertible Neural Network, Central Difference Convolution and Zero-order Filtering  as ``random weights network prototype'' with respect of the following four levels:
	
	\begin{itemize}
		\item the different random weights strategies; 
		
		\item the different network architectures, \emph{eg,} pure convolution layer or transformer; 
		
		\item the different network architecture depths;
		
		\item the different numbers of random weights network combination.  
	\end{itemize}
	
	Based on the above setting, we employ the random weights networks as loss function to better optimize the task model in the following two variants:
	
	\begin{itemize}
		\item the weights are randomly initialized only once during the whole training procedure; 
		
		\item the weights are randomly initialized at each training iteration epoch; 
		
	\end{itemize}

	Our belief is testified over the representative baselines concerning several image restoration tasks including image de-noising, low-light image enhancement and guided image super-resolution and the extensive experimental results demonstrate its effectiveness:
	
	To summarize, we make the following key contributions:
	
	\begin{itemize}
		\item Orthogonal to the existing data and model studies, our proposed paradigm is capable of improving the model performance without changing the original model and data configuration as the baseline;  
		
		\item The proposed random weights network can be plug-and-play into existing deep learning-based image restoration methods and is elegant without any training and testing computational cost; 
		
		\item This is the first attempt to propose the belief ``Random Weights Networks can Be Acted as Loss Prior Constraint for Image Restoration'' and it will spark the realms of loss function not only model and data domain designs. 
		
	\end{itemize}
	
	\textbf{Prophesy.} we are excited about the future of our advocated belief and plan to apply them to other tasks. We plan to extend the proposed belief to problems involving natural language processing and to investigate the underlying  mechanisms. In addition, we believe that our belief will hit the efficient field  by the significant performance, because the violent network architecture designs are not preferred.

	\section{Method}
	\vspace{-0.5mm}
	
	\subsection{Overall Architecture}
	
	In this section, we will first revisit the our belief ``Random Weights Networks can Be Acted as Loss
	Prior Constraint for Image Restoration'' and provide several alternative solutions in the strict Functional manifold to implement our belief, then detail the tailored random weights network over the specific tasks. 
	
	Based on our belief, the corresponding pipeline consists of two steps: 1) firstly randomly initialize the additional loss network in the pre-defined Functional manifold and then 2) employ the random weights network as the loss function to guide the task network learning. In this work, we stand on the shoulder of ``random weights network'' in the strict Functional manifold including Taylor's Unfolding Network, Invertible Neural Network, Central Difference Convolution and Zero-order Filtering  and treat it as the additional loss function network.

	\begin{figure*}[t]
		\setlength{\abovecaptionskip}{-0.3cm}
		\setlength{\belowcaptionskip}{-0.5cm}
		\begin{center}
			\begin{tabular}[t]{c} 
				\includegraphics[width=\textwidth]{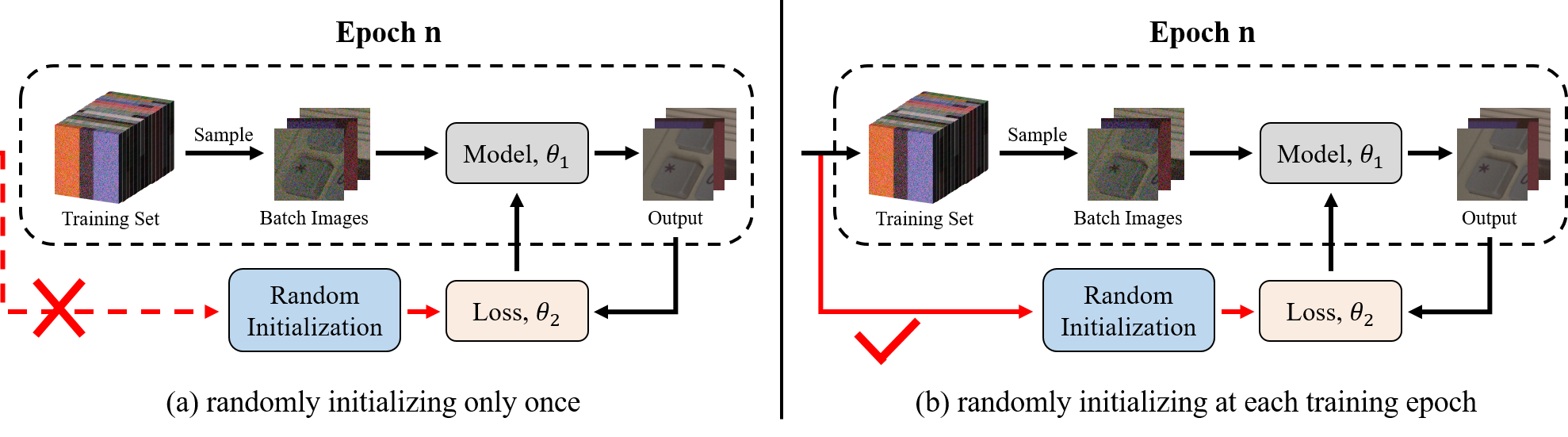}
			\end{tabular}
		\end{center}
		\caption{\small Random Weights Strategies. (a) the weights are randomly initialized only once during the whole training procedure; (b) the weights are randomly initialized at each training iteration epoch; .}
		\label{fig:random}
	\end{figure*}

	\subsection{Image Restoration Flowchart}
	
	Suppose that the task model as $f(\mathbf{x})$ that transforms the input image $\mathbf{x}$ to the output ones $\mathbf{y}$, the process can be written as
	\begin{equation}
		\mathbf{y} = f(\mathbf{x}),
	\end{equation}
	Suppose that the latent ground truth as $\rm GT$,  the original image-level loss function \emph{e.g.,} $\rm L_1$ and $\rm L_2$ is usually employed to take account for optimization, which can be remarked as 
	\begin{equation}
		\mathbf{L} = ||\mathbf{GT}- \mathbf{y}||_{1,2}
	\end{equation}
	where $||.||_{1,2}$ is the image-level loss function \emph{e.g.,} $\rm L_1$ and $\rm L_2$ respectively.  
	
	\textbf{Motivation.} From Bayesian perspective,  it is well known that minimizing $\rm L_1$ and $\rm L_2$  can be equivalent to maximum likelihood estimation in regression. The prediction of a regressor can be considered as the mean of a noisy prediction distribution, which is modeled as a Gaussian or Laplace distribution in the classic probabilistic interpretation as
	\begin{align}
		& L_2: \rm p(y|x;\theta)= \mathcal{N}(GT;y,\sigma_{noise}^{2}I) \\
		& L_1: \rm p(y|x;\theta)= \mathcal{L}(GT;y,b) 
	\end{align}
	where $\sigma_{noise}$ is the scale of an i.i.d. error term $\rm \epsilon \sim \mathcal{N}(0,\sigma_{noise}^{2}I)$ and $\rm b= \sqrt{\frac{\sigma_{noise}}{2}}$ . However, the real distribution is usually complex, thus limiting the model optimization. To this end, we propose to customize the additional network as loss prior regularization term to push the output close to the ground truth distribution.

	\subsection{Taylor's Unfolding Manifold}
	
	\paragraph{Problem Formulation.}
	Inspired by image decomposition, we exploit the Taylor's Unfolding to formulate the decomposition-oriented manifold as
	\begin{equation}
		\label{equ1}
		\boldsymbol y = \boldsymbol{A}\boldsymbol x + \boldsymbol N,
	\end{equation}
	where $\boldsymbol y$, $\boldsymbol x$, $\boldsymbol A$ and $\boldsymbol N$ denote a observation, a latent clear image, a transformation matrix and error term respectively. 
	
	Let $\boldsymbol y_0 = \boldsymbol{A}\boldsymbol x = (\boldsymbol y - \boldsymbol N)$, we learn $\boldsymbol x$ by the function $\boldsymbol F$ 
	\begin{equation}
		\boldsymbol x = \boldsymbol F (\boldsymbol y_0) 
		= \boldsymbol F (\boldsymbol y - \boldsymbol N).
		\label{b3}
		\vspace{-0.6mm}
	\end{equation}
	Let $ -\boldsymbol N  = \boldsymbol \epsilon = \boldsymbol y_0 - \boldsymbol y $, we expand Equation (\ref{b3}) with an infinite-order Taylor's series expansion of the variable $\boldsymbol y$, written compactly as
	\begin{align}
		\boldsymbol x\;
		= {} & \boldsymbol F (\boldsymbol y_0)=  \boldsymbol F(\boldsymbol y\;+\;\boldsymbol\epsilon)\;        \\
		= {} & \boldsymbol F(\boldsymbol y)\!+\!\frac1{1!}\;
		\frac{\partial\boldsymbol F(\boldsymbol y)}{\partial\boldsymbol y}
		{\boldsymbol\epsilon}\!+
		\!\boldsymbol.\boldsymbol.\boldsymbol.
		\!+\!\frac1{k!}\;
		\frac{\partial^k\boldsymbol F(\boldsymbol y)}{\partial^k\boldsymbol y}
		{\boldsymbol\epsilon^k}\!+\;...\;     \label{taylor1}  \\
		= {} & \sum_{k\;=\;0}^\infty\!\frac1{k!}\;
		\frac{\partial^k\boldsymbol F(\boldsymbol y)}{\partial^k\boldsymbol y}
		{\boldsymbol\epsilon^k}\! ,    \label{taylor2}  
		\vspace{-0.8mm}
	\end{align}
	When only regarding $n$ order Taylor's Approximations, it can be simplified as
	\begin{equation}
		\boldsymbol x\;
		= \boldsymbol F(\boldsymbol y)\;+\;\sum_{k\;=\;1}^n \frac{\partial^k\boldsymbol F(\boldsymbol y)}{\partial^k\boldsymbol y}
		{\boldsymbol\epsilon^k}\!.    \label{n_taylor}
		\vspace{-0.6mm}
	\end{equation}
	It can be separated into two parts for consideration. The first term, defining as main part, $\boldsymbol F(\boldsymbol y)$ represents the high-level contextualized information while the rest is the local high-order spatial details as the goal of image decomposition. 
	
	\textbf{Taylor's Manifold Implementation.}
	Based on above Taylor’s Unfolding Formula as blueprints, Taylor's Manifold consists of two operation parts: \textbf{Mapping Function Part $\boldsymbol F$} and \textbf{Derivative Function Part $\boldsymbol G$}. 
	
\begin{figure}[t]
		\centering
		\includegraphics[width=\columnwidth]{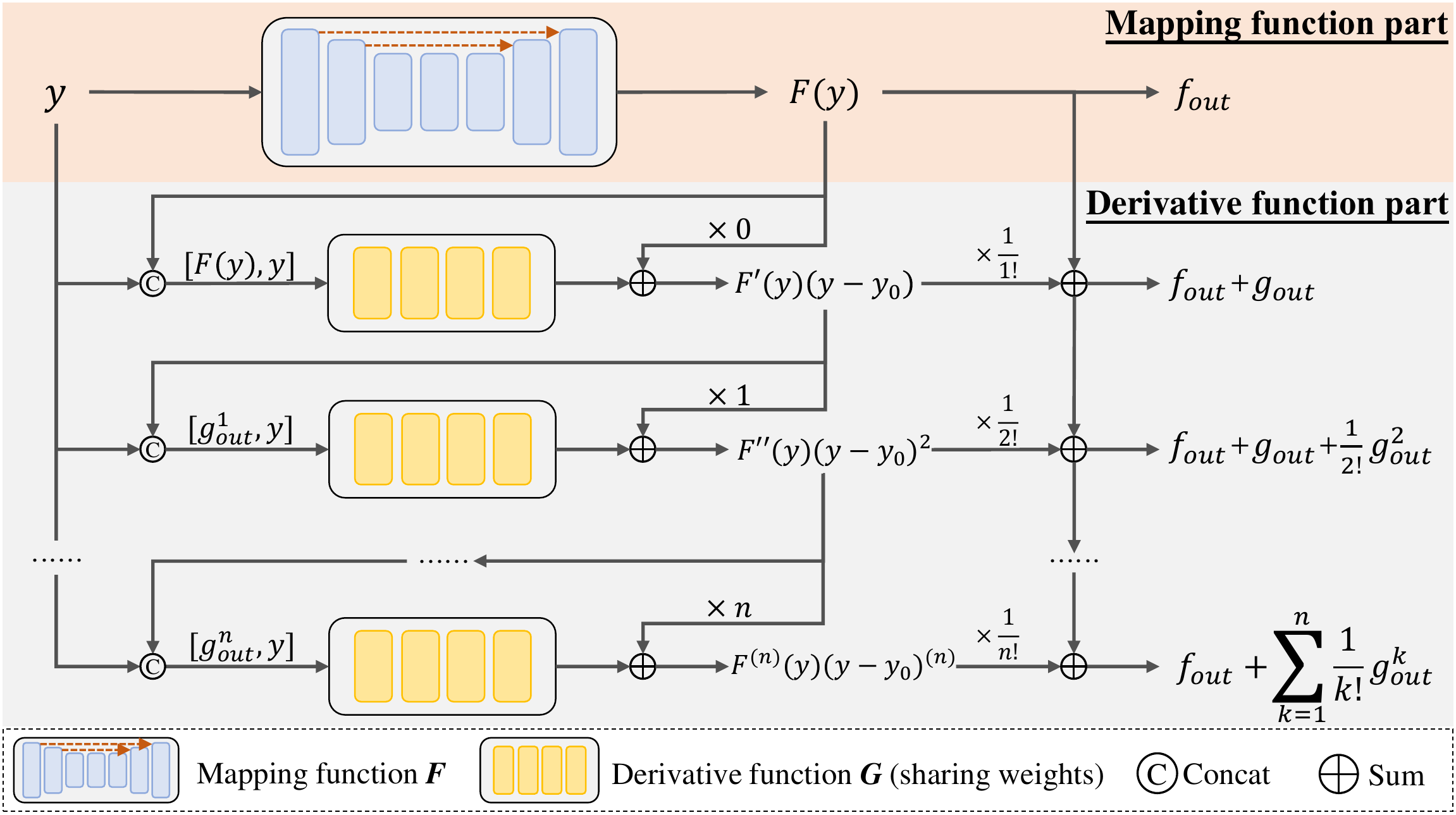}
		\caption{The overview structure of Deep Taylor's Approximations Framework. It consists of two parts, Mapping function part $\boldsymbol F$ and Derivative function part $\boldsymbol G$.  The parameters of Derivative function part $\boldsymbol G$ are shared across the progressive stages.
		}
		\label{fig:taylor}
		\vspace{-2mm}
	\end{figure}

	\textbf{Derivative Function Part $\boldsymbol G$.}
	Recalling Equation (\ref{n_taylor}), for the $k$ order derivative part, it can be written as
	\begin{equation}
		{\boldsymbol F}^{\boldsymbol(k \boldsymbol)}\boldsymbol(\boldsymbol y \boldsymbol)\boldsymbol({\boldsymbol \epsilon}\boldsymbol)^k 
		\boldsymbol = \frac{\partial^k\boldsymbol F(\boldsymbol y)}{\partial^k\boldsymbol y}
		{\boldsymbol\epsilon^k}\!,
	\end{equation}
	differentiating above $k$ order part ${\boldsymbol F}^{\boldsymbol(k \boldsymbol)}\boldsymbol(\boldsymbol y \boldsymbol)\boldsymbol({\boldsymbol \epsilon}\boldsymbol)^k$ for $y$ as
	\begin{equation}
		\frac{\partial {\boldsymbol F}^{\boldsymbol(k \boldsymbol)}\boldsymbol(\boldsymbol y \boldsymbol)\boldsymbol({\boldsymbol \epsilon}\boldsymbol)^k}
		{\partial \boldsymbol y}\;
		= \;{\boldsymbol F}^{\boldsymbol({k+1} \boldsymbol)}\boldsymbol(\boldsymbol y \boldsymbol)\boldsymbol({\boldsymbol \epsilon}\boldsymbol)^k
		\;\;-\;\;{\boldsymbol F}^{\boldsymbol(k \boldsymbol)}\boldsymbol(\boldsymbol y \boldsymbol) \times k\boldsymbol({\boldsymbol \epsilon}\boldsymbol)^{k-1},
		\;\;  
		\label{grad}
	\end{equation}
	multiplying above Equation (\ref{grad}) by $\epsilon$ as
	\begin{align}
		& \frac{\partial {\boldsymbol F}^{\boldsymbol(k \boldsymbol)}\boldsymbol(\boldsymbol y \boldsymbol)\boldsymbol({\boldsymbol \epsilon}\boldsymbol)^k}
		{\partial \boldsymbol y}\;\times\;\boldsymbol \epsilon\; \\\;
		= &\;({\boldsymbol F}^{\boldsymbol({k+1} \boldsymbol)}\boldsymbol(\boldsymbol y \boldsymbol)\boldsymbol({\boldsymbol \epsilon}\boldsymbol)^k
		\;\;-\;\;k{\boldsymbol F}^{\boldsymbol(k \boldsymbol)}\boldsymbol(\boldsymbol y \boldsymbol)\boldsymbol({\boldsymbol \epsilon}\boldsymbol)^{k-1})
		\;\;\times\;\boldsymbol \epsilon\;\\\;
		=&\;{\boldsymbol F}^{\boldsymbol({k+1} \boldsymbol)}\boldsymbol(\boldsymbol y \boldsymbol)\boldsymbol({\boldsymbol \epsilon}\boldsymbol)^{k+1}
		\;\;-\;\;k{\boldsymbol F}^{\boldsymbol(k \boldsymbol)}\boldsymbol(\boldsymbol y \boldsymbol)\boldsymbol({\boldsymbol \epsilon}\boldsymbol)^k. \label{g1}
	\end{align}
	To this end, we exploit Derivative function sub-network, named $\boldsymbol G$ to take effect as above process. We denote the $k$ order output of network $\boldsymbol G$ as ${\boldsymbol F}^{\boldsymbol(k \boldsymbol)}\boldsymbol(\boldsymbol y \boldsymbol)\boldsymbol({\boldsymbol \epsilon}\boldsymbol)^k$, simply recorded as $ g_{out}^k $. Referring Equation (\ref{g1}), we can find out the connection between $k$ order output and $k+1$ order one
	\begin{equation}
		g_{out}^{k+1} = \boldsymbol G(g_{out}^{k}) + k{\boldsymbol F}^{\boldsymbol(k \boldsymbol)}\boldsymbol(\boldsymbol y \boldsymbol)\boldsymbol({\boldsymbol \epsilon}\boldsymbol)^k.
	\end{equation}
	Replacing ${\boldsymbol F}^{\boldsymbol(k \boldsymbol)}\boldsymbol(\boldsymbol y \boldsymbol)\boldsymbol({\boldsymbol \epsilon}\boldsymbol)^k$ with $ g_{out}^k $ as
	\begin{equation}
		g_{out}^{k+1} = \boldsymbol G (g_{out}^{k}) + k \cdot g_{out}^{k}.
		\label{h1}
	\end{equation}
	Referring above analysis, Taylor's Unfolding Manifold is detailed in Figure \ref{fig:taylor} where the Mapping function part $\boldsymbol F$  maps input as the main energy
	\begin{equation}
		f_{out}\;=\;\boldsymbol F({\boldsymbol y}),
	\end{equation}
	referring Equation (\ref{h1}), the Derivative function part  $\boldsymbol G$ needs the $g_{out}^{k}$ from the Mapping function sub-network $\boldsymbol F$ and the input image $y$. It is because that the unfolding iteration process of $\boldsymbol G$ running involves $y$. In this regard, concatenating  $g_{out}^{k}$ and $y$ into  $\boldsymbol G$ as input for inference
	\begin{equation}
		g_{out}^{k+1}\;=\;\boldsymbol G(Concat(\lbrack g_{out}^k,\;{\boldsymbol y}\rbrack)).
	\end{equation}
	Taking together above two operation steps, the final output of $n$ order Deep Taylor's Approximations framework can be obtained as
	\begin{equation}
		O\;=\;f_{out}\;+\;\sum_{k\;=\;1}^n {\frac1{k!} g_{out}^k}.
	\end{equation}

	\subsection{Invertible Neural Network Manifold}
	\textbf{Problem Formulation.}
	Inspired by image transformation \emph{e.g., Fourier transform and wavelet transform}, we can remark the above invertible transformation  as
	\begin{equation}
		\label{equ1}
		\boldsymbol y = \boldsymbol{A}\boldsymbol x,
		\vspace{-2mm}
	\end{equation}
	where $A$ is the wavelet function for wavelet transform while $A$ transfers to Trigonometric basis for Fourier transform.
	
\begin{figure}[t]
		\setlength{\abovecaptionskip}{-0.2cm}
		\setlength{\belowcaptionskip}{-0.2cm}
		\centering
		\includegraphics[width=0.88\columnwidth]{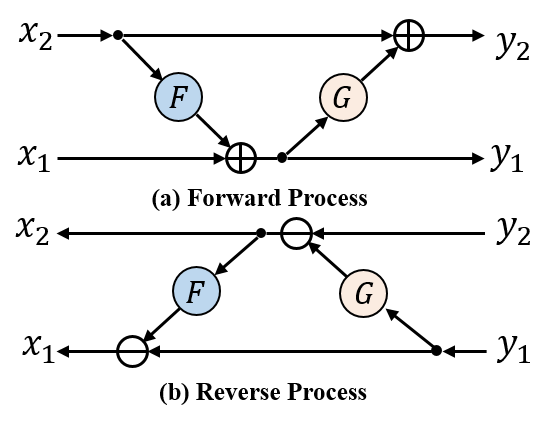}
		\caption{(a) the forward, and (b) the reverse computation of invertible neural network.
		}
		\label{fig:inn}
	\end{figure}

	To model the general invertible manifold, invertible Neural Network is tailored. The basic invertible units partition the input into two groups, denoted $x_1$ and $x_2$ by the channel dimension.
	Each reversible block takes inputs $(x_1, x_2)$ and produces outputs $(y_1, y_2)$ according to the following additive coupling rules -- inspired by NICE's \citep{IRN,realNVP,NICE}
	transformation as shown in Figure \ref{fig:inn}:
	\begin{align}
		y_1 &= x_1 + \mathcal{F}(x_2) \nonumber \\
		y_2 &= x_2 + \mathcal{G}(y_1) 
		\label{innforward}
	\end{align}
	Each layer's activations can be reconstructed from the next layer's activations as follows:
	\begin{align}
		x_2 &= y_2 - \mathcal{G}(y_1) \nonumber \\
		x_1 &= y_1 - \mathcal{F}(x_2)
	\end{align}
	Note that the translation functions $F$ and $G$ are not necessarily invertible and the basic invertible unit is always invertible. The theoretical proof is remarked as bellow.
	
	\textbf{Theoretical proof.} To calculate the Jacobian matrix, the  coding formula (Eq. \ref{innforward}) is more intuitively written as
	\begin{align}
		y_1 &= x_1 + \mathcal{F}(x_2) \nonumber \\
		y_2 &= x_2 + \mathcal{G}(x_1 + \mathcal{F}(x_2))
	\end{align}
	
	Its Jacobian matrix is 
	
	\begin{align}
		J_{f} & = \left[\begin{array}{ll}
			\frac{\partial y_{1}}{\partial x_{1}} & \frac{\partial y_{1}}{\partial x_{2}} \\
			\frac{\partial y_{2}}{\partial x_{1}} & \frac{\partial y_{2}}{\partial x_{2}}
		\end{array}\right] & = \left[\begin{array}{cc}
			1 & \frac{\partial F}{\partial x_{2}} \\
			\frac{\partial G}{\partial x_{1}} & 1+\frac{\partial G}{\partial F} \frac{\partial F}{\partial x_{2}}
		\end{array}\right]
	\end{align}
	Due to the fact 
	$\frac{\partial G}{\partial x_1} = \frac{\partial G}{\partial F}$, we thus calculate the above Jacobian matrix as
	\begin{align}
		J_f &= 1\times (1+ \frac{\partial G}{\partial F}\frac{\partial F}{\partial x_2})-\frac{\partial G}{\partial x_1}\frac{\partial F}{\partial x_2}\\ \nonumber
		&= 1 + \frac{\partial F}{\partial x_2}(\frac{\partial G}{\partial x_1} - \frac{\partial G}{\partial F}) = 1-\frac{\partial F}{\partial x_2} \times 0 \\ \nonumber
		&=1
	\end{align}
	
	\subsection{Central Difference Convolution Manifold}
	As the basic operator in deep networks, the vanilla 2D convolution consists of two main steps: 1) \textsl{sampling} local neighbor region $\mathcal{R}$ over the input feature map $x$; and then 2) \textsl{aggregating} the sampled values via learnable weights $w$. As a result, the output feature map $y$ can be formulated as
	\begin{equation} 
		y(p_0)=\sum_{p_n\in \mathcal{R}}w(p_n)\cdot x(p_0+p_n),
		\label{eq:vanilla}
	\end{equation}
	where $p_0$ denotes the current location on both input and output feature maps while $p_n$ enumerates the locations in $\mathcal{R}$. For instance, local receptive field region for convolution operator with 3$\times$3 kernel and dilation 1 is $\mathcal{R}=\left \{  (-1,-1),(-1,0),\cdots,(0,1),(1,1)  \right \}$. 
	\begin{figure}
		\centering
		\includegraphics[width=0.9\columnwidth]{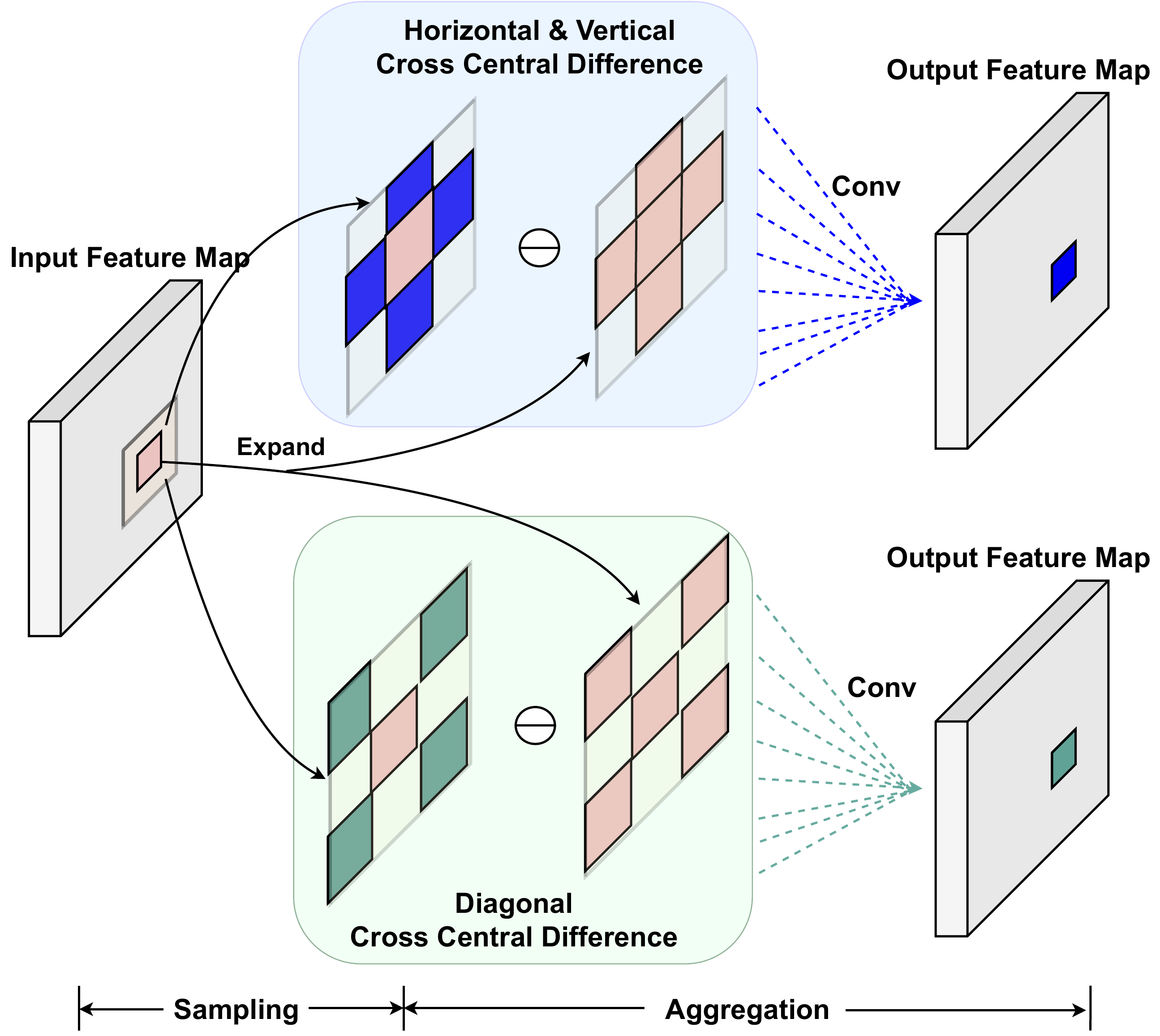}
		\caption{\small{
				Cross central difference convolution. The C-CDC(HV) in the upper part calculates the central gradients from the horizontal \& vertical neighbors while the C-CDC(DG) in the lower part from the diagonal neighbors.
			}
		}
		\label{fig:CCDC}
		\vspace{-1.4em}
	\end{figure}
	
	Different from the vanilla convolution, the CDC introduces central gradient features to enhance the representation and generalization capacity, which can be formulated as
	\begin{equation} 
		y(p_0)=\sum_{p_n\in \mathcal{R}}w(p_n)\cdot (x(p_0+p_n)-x(p_0)).
		\label{eq:CCDC}
	\end{equation}
	
	In Figure \ref{fig:CCDC}, the generalized CDC operator can be represented by combination of vanilla convoluiton and CDC
	\begin{equation} 
		\begin{split}
			&y(p_0)
			=\theta \cdot \underbrace{\sum_{p_n\in \mathcal{R}}w(p_n)\cdot (x(p_0+p_n)-x(p_0))}_{\text{central difference convolution}}\\
			&\quad+ (1-\theta)\cdot \underbrace{\sum_{p_n\in \mathcal{R}}w(p_n)\cdot x(p_0+p_n)}_{\text{vanilla convolution}}, \\
			&= \underbrace{\sum_{p_n\in \mathcal{R}}w(p_n)\cdot x(p_0+p_n)}_{\text{vanilla convolution}}+\theta\cdot (\underbrace{-x(p_0)\cdot\sum_{p_n\in \mathcal{R}}w(p_n))}_{\text{central difference term}},\\
		\end{split}
		\label{eq:CCDC2}
	\end{equation}
	where hyperparameter $\theta \in [0,1]$ trade-offs the contribution between intensity-level and gradient-level information.

	\subsection{Reverse Filtering Network Manifold}
	
	\paragraph{Definition 3.2} \label{def} \textit{Suppose $(\mathcal{H},d)$ is a metric space and $T:\mathcal{H} \rightarrow \mathcal{H}$ is a mapping function. For all $x, y \in \mathcal{H}$, if there exists a constant $c \in [0,1)$ that makes the following formula
		\begin{equation}
			d(T(x), T(y)) \leq cd(x, y),
		\end{equation}
		mapping $T:\mathcal{H} \rightarrow \mathcal{H}$ is called Contraction Mapping}. 
	
	\paragraph{Theorem 3.2} \label{Theorem} \textit{A variable $x^{*}$ is a fixed point for a given function $\Phi$ if $\Phi(x^{*})=x^{*}$. When mapping  $\Phi:\mathcal{H} \rightarrow \mathcal{H}$ is a contraction mapping, $\Phi$ admits a unique fixed-point $x^{*}$ in $\mathcal{H}$. Further, $x^{*}$ can be found in the following way. Let the initial guess be $x_{0}$ and define a sequence $\left\{x_{n}\right\}$ as $x_{n}=\Phi(x_{n-1})$. When the iterative process converges, $\lim _{n \rightarrow \infty} x_{n}=x^{*}$. } 
	
	\textbf{Reverse Filtering.}   $f(\cdot)$ can be considered as broadly defined filters that smooth images and filtering process can be described as
	$\boldsymbol{y}=f(\boldsymbol{x}).$
	where $\boldsymbol{x}$ and $\boldsymbol{y}$ are the input image and the filtering result. The reverse filtering can estimate $\boldsymbol{x}$ without needing to compute $f^{-1}(\cdot)$ and update restored images according to the filtering effect as
	\begin{equation} 
		\boldsymbol{x}^{k+1}=\boldsymbol{x}^{k}+\boldsymbol{y}-f(\boldsymbol{x}^{k}),
	\end{equation}
	where $\boldsymbol{x}^{k}$ is the current estimate of $\boldsymbol{x}$ in the \textit{k}-th iteration. The iteration starts from $\boldsymbol{x}^{0}=\boldsymbol{y}$ and $\boldsymbol{x}^{k}$ gets closer and closer to $\boldsymbol{x}$ with the increasing ${k}$. We make auxiliary function $\varphi(\cdot)$ as
	$ \label{eq:iteration}
	\varphi(\boldsymbol{x})=\boldsymbol{x}+\boldsymbol{y}-F(\boldsymbol{x}).
	$
	Therefore, the above iterative process can be regarded as a fixed point iteration 
	\begin{equation}
		\boldsymbol{x}^{k+1}=\varphi(\boldsymbol{x}^{k}).
	\end{equation}

\begin{figure}[t]
		\centering
		\includegraphics[width=\columnwidth]{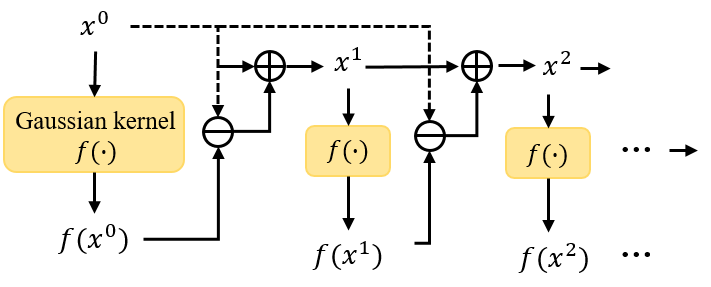}
		\caption{The reverse filtering structure.
		}
		\label{fig:zero}
		\vspace{-3mm}
	\end{figure}

	\textbf{Reverse Filtering Network Manifold.} \label{Network}
	In Figure \ref{fig:zero}, the filtering process $f$ is implemented by \textbf{Multi-scale Gaussian Convolution Module} and thus satisfies the sufficient condition for definition ~\ref{def}. Take the $\varphi_{1}(\cdot)$ for instance, the sufficient condition that theorem ~\ref{Theorem} holds is that $\varphi_{1}(\mathbf{H})$ forms a contraction mapping
	\begin{eqnarray} \label{in-equation}
		\begin{aligned}
			&\left\|\varphi_{1}\left(\mathbf{H}_{a}\right)-\varphi_{1}\left(\mathbf{H}_{b}\right)\right\| \\&=\left\|\left[\mathbf{H}_{a}+\mathbf{\hat{L}}-f\left(\mathbf{H}_{a}\right)\right]-\left[\mathbf{H}_{b}+\mathbf{\hat{L}}-f\left(\mathbf{H}_{b}\right)\right]\right\|  \\&=\left\|\left[\mathbf{H}_{a}-f\left(\mathbf{H}_{a}\right)\right]-\left[\mathbf{H}_{b}-f\left(\mathbf{H}_{b}\right)\right]\right\|  \leq c \cdot\left\|\mathbf{H}_{a}-\mathbf{H}_{b}\right\|
		\end{aligned}
	\end{eqnarray}
	For linear filters, the condition is further simplified as
	\begin{equation}
		\|\mathbf{H}-f(\mathbf{H})\| \leq c \cdot\|\mathbf{H}\|. \quad c \in[0,1)
	\end{equation}
	
	\subsection{Random Weights Strategies}
	Based on the above setting, as shown in Figure \ref{fig:random}, we employ the random weights networks as loss function to better optimize the task model in the following two variants:
	
	\begin{itemize}
		\item the weights are randomly initialized only once during the whole training procedure; 
		
		\item the weights are randomly initialized at each training iteration epoch, denoted as ``\textbf{epochR}''; 
		
	\end{itemize}

	\subsection{Pipeline}
	Suppose that the randomly initialized manifold model as $f_{random}(.)$, it is employed as the complementary loss function to the original image-level loss function \emph{e.g.,} $\rm L_1$ and $\rm L_2$. The total loss function is remarked as 
	\begin{equation}
		\mathbf{L} = ||\mathbf{GT}- \mathbf{y}||_{1,2} + \lambda ||f_{random}(\mathbf{GT})-f_{random}(\mathbf{y})||_{1,2}
	\end{equation}
	where $\lambda$ indicates the weighted factor, $||.||_{1,2}$ is the image-level loss function \emph{e.g.,} $\rm L_1$ and $\rm L_2$ and $\mathbf{GT}$ denotes the ground truth.  
	
	\section{Experiments}
	\vspace{-1.2mm}
	To demonstrate the efficacy of our proposed belief, we conduct extensive experiments on multiple image restoration tasks, including image de-noising, low-light image enhancement, and guided image super-resolution. \textbf{\textit{We provide more experimental results in the Appendix.}}

	\begin{table}[!htb]
		\small
		\centering
		\renewcommand{\arraystretch}{1.2}
		\caption{\textbf{Quantitative comparisons of image enhancement.}}
		\begin{tabular}{l l |c c c c c c }
			\hline
			\multirow{2}{*}{Model} & \multirow{2}{*}{Configurations}& \multicolumn{3}{c}{LoL}  \\
			&&PSNR & SSIM & NIQE \\
			\hline
			\multirow{5}{*}{SID} & Original& 20.2461 & 0.7920 & 4.1586  \\ \hline
			
			&+Taylor & 20.5864 & 0.7971 & 3.8348  \\
			&+Taylor+epochR & 20.6018 & 0.7975 & 3.8079  \\

			&+CDC  & 20.3298& 0.7927 & 3.7431  \\
			&+CDC+epochR  & 20.4750 & 0.7999& 3.6636  \\

			&+INN  & 20.3178 & 0.7944 & 3.8889   \\
			&+INN+epochR  & 20.3958 & 0.7924 & 3.9210   \\

			&+Reverse & 20.5014 & 0.7941 & 4.0841   \\
			&+Reverse+epochR & 20.5203 & 0.7943 & 4.0654  \\

			\hline

			\multirow{5}{*}{DRBN} & Original& 19.8509  & 0.7769 & 4.7738 \\ \hline

			&+Taylor & 20.1156 & 0.7778 & 4.6767 \\
			&+Taylor+epochR & 20.2405 & 0.7791 & 4.6721 \\

			&+CDC  & 19.7952 & 0.7851 & 4.7886 \\
			&+CDC+epochR  & 20.0756 & 0.7837 & 4.7850 \\

			&+INN  & 19.8543 & 0.7774 & 4.6741 \\
			&+INN+epochR  & 20.1913 & 0.7769 & 4.8067 \\

			&+Reverse & 19.9547 & 0.7765 & 4.6265 \\
			&+Reverse+epochR & 20.1358 & 0.7751 & 4.7716 \\

			\hline
		\end{tabular}
		\label{image enhancement}
		\vspace{-2mm}
	\end{table}

	\begin{table}[!t]
		\centering
		\caption{\textbf{Quantitative comparisons for image de-noising.}}
		
		\begin{tabular}{l l |c c  c c  }
			\hline
			\multirow{2}{*}{Model} & \multirow{2}{*}{Configurations}& \multicolumn{2}{c}{SIDD}  \\
			&&PSNR$\uparrow$ & SSIM$\uparrow$   \\
			\hline
			\multirow{5}{*}{DnCNN} & Original& 37.1992 & 0.8954   \\ \hline
			&+Taylor & 37.3163 & 0.8955   \\ 
			&+Taylor+epochR & 37.3719  & 0.8954  \\

			&+CDC  & 37.2329 & 0.8958  \\
			&+CDC+epochR  & 37.2784  & 0.8955 \\

			&+INN  & 37.3168 & 0.8970   \\
			&+INN+epochR  & 37.3318 & 0.8964  \\

			&+Reverse & 37.3162 & 0.8965   \\
			&+Reverse+epochR & 37.3321 & 0.8955  \\

			\hline
			\multirow{5}{*}{MPRnet} & Original& 39.2372
			& 0.9159  \\ \hline
			&+Taylor & 39.2953 & 0.9161   \\
			&+Taylor+epochR & 39.3283  & 0.9161  \\

			&+CDC  & 39.2609 & 0.9160   \\
			&+CDC+epochR  & 39.2821  & 0.9161  \\

			&+INN  & 39.2729 & 0.9162  \\
			&+INN+epochR  & 39.3317 & 0.9162  \\

			&+Reverse & 39.2446 & 0.9160   \\
			&+Reverse+epochR & 39.2660 & 0.9161   \\
			
			\hline
		\end{tabular}
		\label{image de-noising}
		\vspace{-5mm}
	\end{table}

	\begin{table*}[!t]
		\centering
		\caption{\textbf{Quantitative comparisons of guided image super-resolution.}}
		\resizebox{0.9\linewidth}{!}{ 
			\begin{tabular}{l l |c c c c  c c c c}
				\hline
				\multirow{2}{*}{Model} & \multirow{2}{*}{Configurations}& \multicolumn{4}{c}{WorldView-II} & \multicolumn{4}{c}{GaoFen2} \\
				& &PSNR$\uparrow$ & SSIM$\uparrow$&SAM$\downarrow$&ERGAS$\downarrow$ &PSNR$\uparrow$ & SSIM$\uparrow$&SAM$\downarrow$  &EGAS$\downarrow$ \\
				\hline
				\multirow{3}{*}{INNformer} & Original& 41.6903 & 0.9704 & 0.0227 & 0.9514         & 47.3528 & 0.9893 & 0.0102 & 0.5479  \\ \hline
				
				&  +Taylor & 41.8168 & 0.9716 & 0.0224 & 0.9276  & 47.4058 & 0.9901 & 0.0101 & 0.5356   \\
				
				&+CDC & 41.8072 & 0.9715 & 0.0224 & 0.9276  & 47.4121 & 0.9902 & 0.0100 & 0.5354  \\
				
				&+INN & 41.8229 & 0.9717 & 0.0223 & 0.9276  & 47.4233 & 0.9904 & 0.0100 & 0.5353  \\
				
				&+Reverse & 41.7293 & 0.9711 & 0.0226 & 0.9276  & 47.4010 & 0.9901 & 0.0101 & 0.5354  \\
				
				\hline
				
				\multirow{3}{*}{SFINet} &  Original& 41.7244 & 0.9725 & 0.0220 & 0.9506       & 47.4712 & 0.9901 & 0.0102 & 0.5462 \\ \hline
				
				& +Taylor & 41.9314 & 0.9723 & 0.0219 & 0.9278    & 47.6132 & 0.9911 & 0.0101 & 0.5277  \\
				
				& +CDC & 41.8943 & 0.9719 & 0.0220 & 0.9283     & 47.5990 & 0.9910 & 0.0101 & 0.5281  \\
				
				& +INN & 41.9521 & 0.9727 & 0.0217 & 0.9278     & 47.6316 & 0.9916 & 0.0101 & 0.5275  \\
				
				& +Reverse & 41.9217 & 0.9722 & 0.0218 & 0.9281     & 47.6227 & 0.9914 & 0.0101 & 0.5275  \\
				\hline
		\end{tabular}}
		\label{tab:ps}
		\vspace{-5mm}
	\end{table*}
	
	\subsection{Experimental Settings} \label{4.1section}
	\textbf{Image enhancement.} We verify our belief on the image enhancement benchmarks, LOL \cite{Chen2018Retinex}. Further, we adopt the two promising baselines, SID \cite{chen2018learning} and DRBN \cite{yang2020fidelity}.

	\textbf{Image De-noising.}  Following \cite{9577298}, to evaluate our belief on the image de-noising task, we employ the widely-used SIDD dataset as training benchmark. Further, the corresponding performance evaluation is conducted on the remaining validation samples from the SIDD dataset \cite{abdelhamed2018high}. Two representative image de-noising algorithms DnCNN \cite{zhang2017beyond} and MPRnet \cite{9577298} are selected as the baselines.
	
	\textbf{Guided Image Super-resolution.} Following \cite{9662053,sfinet}, we adopt the pan-sharpening, the representative task of guided image super-resolution, for evaluations. The WorldView II and GaoFen2 datasets  \cite{9662053,sfinet} are used for experimental implementations. We employ  two state-of-the-art methods including  INNformer \cite{9662053} and SFINet \cite{sfinet} as the baselines.

	\subsection{Implementation details} \label{4.2section}
	
	For simplification, we denote some annotations of the proposed alternative solutions in strict mathematical manifolds before our presentation and  the implementation variants of the baselines are organized as the five configurations:
	\begin{itemize}
		\setlength{\itemsep}{0pt}
		\setlength{\parsep}{0pt}
		\setlength{\parskip}{0pt}
		\vspace{-2mm}
		\item [1)] \textbf{Original}: the baseline with the original image-level loss function ($\rm L_1$ or $\rm L_2$); 
		
		\item [2)] \textbf{+Taylor}: complementing the original image-level loss function  ($\rm L_1$ or $\rm L_2$)  with the Taylor's unfolding network manifold, forming the total loss;
		
		\item[3)] \textbf{+CDC}: complementing the original image-level loss function  ($\rm L_1$ or $\rm L_2$)  with the central difference convolution manifold as the total loss; 
		
		\item[4)] \textbf{+INN}:  complementing the original image-level loss   with the invertible neural network manifold;
		
		\item[5)] \textbf{+Reverse}:   complementing the original image-level loss   with the reverse filtering network manifold.

	\end{itemize}

	\subsection{Comparison and Analysis} \label{integration}
	
	\textbf{Quantitative Comparison.} We perform the model performance comparison over different configurations, as described in implementation details. The quantitative results of image de-noising, low-light image enhancement and guided image super-resolution are presented in Table \ref{image de-noising},  Table \ref{image enhancement} and Table \ref{tab:ps} where the best are highlighted in bold. From the results, by integrating with our proposed belief of the alternative manifolds, we can observe performance gain against the baselines across all the datasets in all tested tasks, suggesting the effectiveness of our approach.  Specifically, in terms of image enhancement, the baseline DRBN with ``+Taylor'', ``+CDC'', ``INN'' and ``Reverse'' has obtained the 0.4dB, 0.3dB, 0.2dB, 0.2dB PSNR gains.  

	\textbf{Qualitative  Comparison.} Due to the limited space, we report the visual results of the de-noising/enhancement task in the Appendix that can more clearly show the effectiveness of our proposed belief. As shown in Figure \ref{dndn} to Figure \ref{ie}, integrating the proposed belief with the original baseline achieves more visually pleasing results.

	\begin{figure}[t]
		\centering
		\includegraphics[width=0.9\columnwidth]{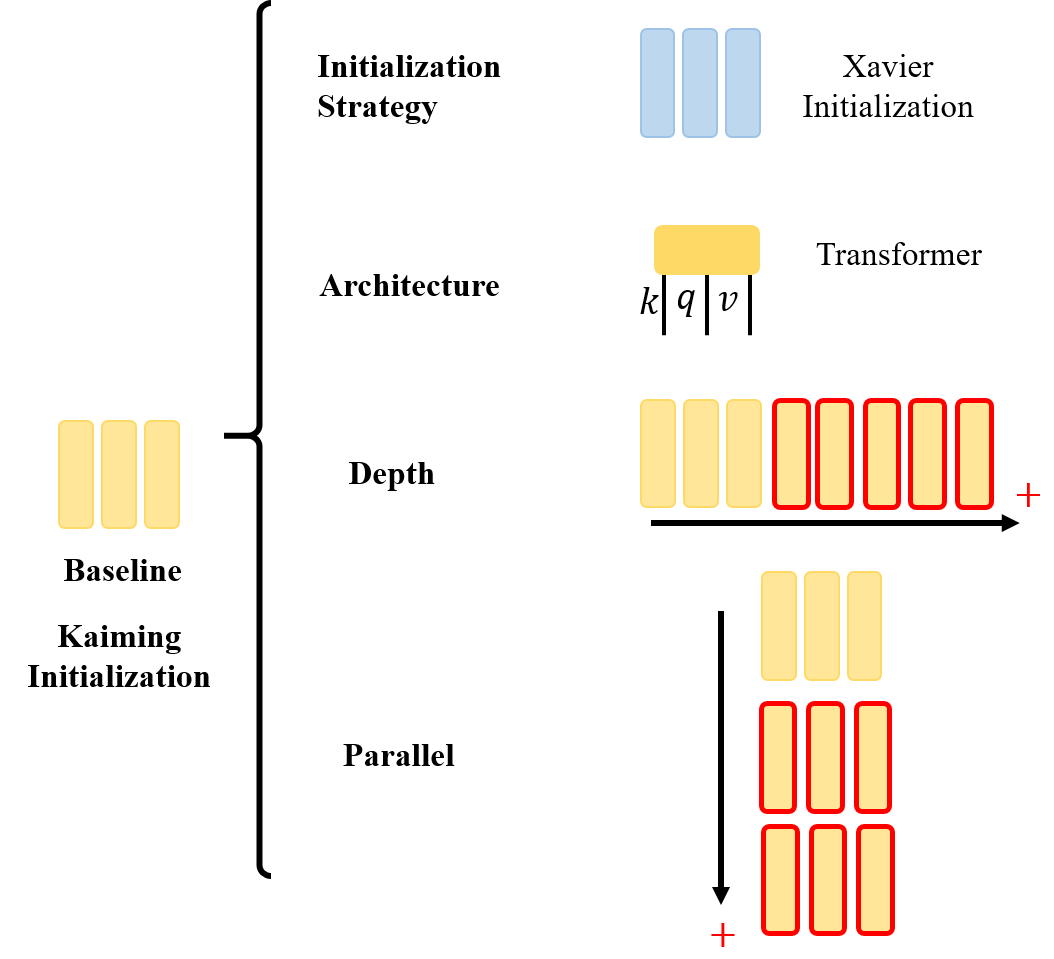}
		\caption{The ablation studies: 1) initialization strategy, 2) model architecture, 3) model depth and 4) model numbers.
		}
		\label{fig:ablation}
		\vspace{-5mm}
	\end{figure}
	
	\vspace{-3mm}
	\section{Ablation Studies}
	To verify the stability of our belief, we conduct the following variants with respect of the following four levels: 1) initialization strategy, 2) model architecture, 3) model depth and 4) model numbers. \textbf{\textit{The experimental results of ablation studies are presented in the Appendix.}}
	
	\textbf{Initialization strategy.} In our work, the default initialization strategy is Kaiming initialization. To explore the impact of initial mode,  we replace the default Kaiming initialization by Xavier initialization. Table \ref{image de-noising as init} and Table \ref{image enhancement as init} show that replacing the default almost has little impact on performance, thus verifying the robustness of our belief.
	
	\textbf{Model architecture.} All of the loss networks are implemented by convolution network as default. To explore the architecture impact,  we replace the default CNN by Transformer. The results in Table \ref{image de-noising as arc} and Table \ref{image enhancement as arc} demonstrate that replacing it rarely affects the performance. 
	
	\textbf{Model depth.}  For model depth, we change the model depth of loss network by adding the layers. To ensure a fair comparison, the other factors keep the same. The  results in Table \ref{image de-noising as depth} and Table \ref{image enhancement as depth} demonstrate the stable performance.
	
	\textbf{Model numbers.} In our experiment, we use the single loss network as default. As shown in Table \ref{image de-noising as num} and Table \ref{image enhancement as num}, we employ multiple parallel loss networks  to verify the impact of model numbers. The results indicates that increasing the number of models will improve the performance. It attributes to the advantages of model ensemble.
	
	\vspace{-3mm}
	\section{Conclusion}
	\vspace{-1mm}
	In this paper, we resort our efforts to investigate the potential of loss function and present our belief \textbf{``Random Weights Networks can Be Acted as Loss Prior Constraint for Image Restoration''}. Inspired by Functional theory, we provide several alternative solutions in the strict mathematical manifolds  as ``random weights network prototype''.  Our proposed belief can be directly plugged into existing image restoration networks and extensive experiments demonstrate its effectiveness of our belief.

	\nocite{langley00}
	
	\bibliography{example_paper}
	\bibliographystyle{icml2023}

	\newpage
	\appendix
	\onecolumn
	\section{\textbf{Appendix}.}
	\subsection{Quantitative comparison.}
	
	\textbf{Guided Image super-resolution.}  The quantitative results for pan-sharpening are summarized in Tables \ref{tab:ps} where the best results are highlighted in bold. From the results, by integrating with our proposed random weights network by alternative mathematical  manifolds, all the reported baselines have  achieved consistent performance gains across all the datasets in terms of  all metrics, suggesting the effectiveness of our belief. 
	
	\subsection{Visual comparison.}
\begin{figure}[t]
		\centering
		\includegraphics[width=\columnwidth]{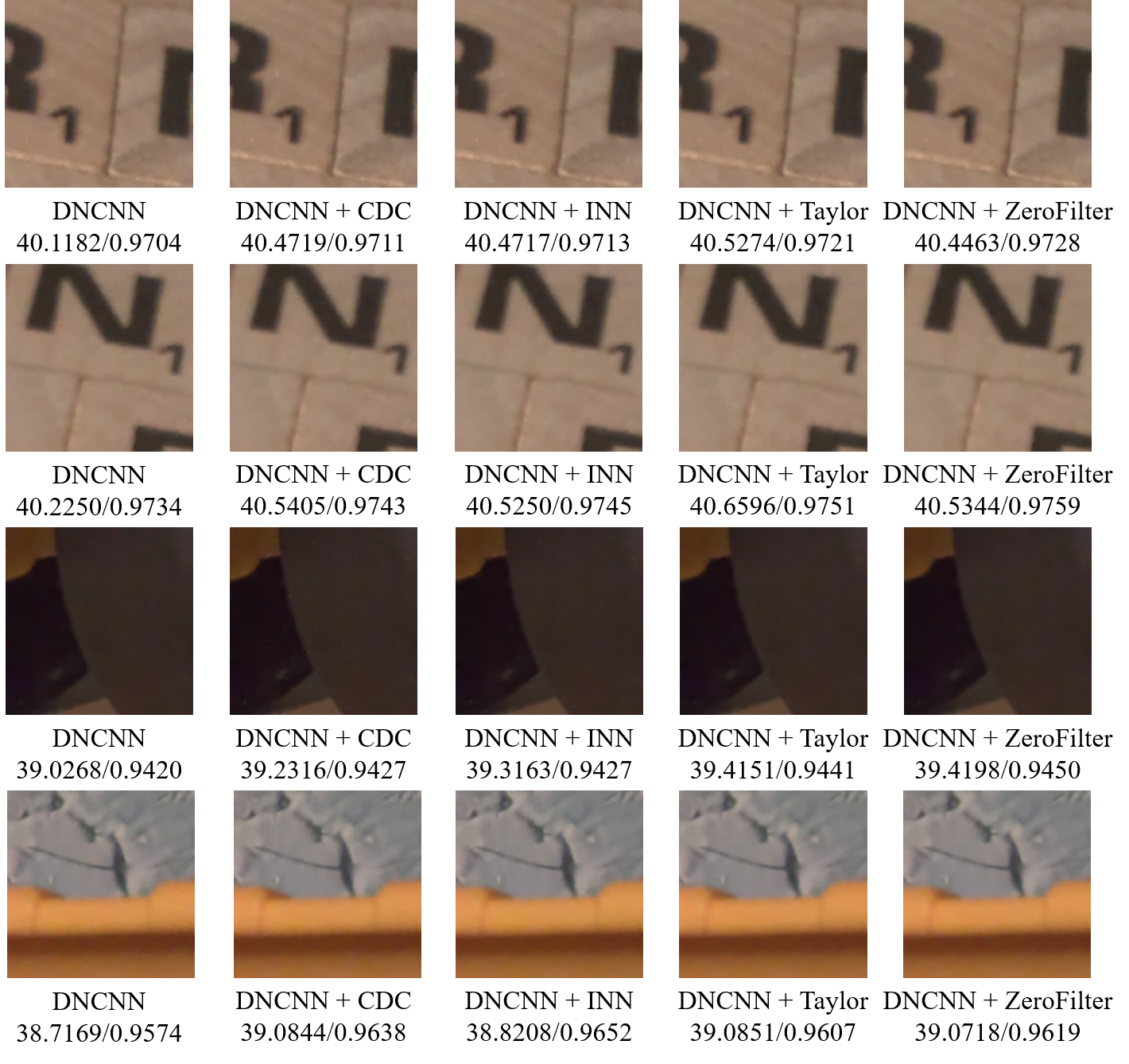}
		\caption{The visual comparison for the image de-noising. We also list the PSNR/SSIM scores under each case.}
		\label{dndn}
	\end{figure}

\begin{figure}[t]
		\centering
		\includegraphics[width=\columnwidth]{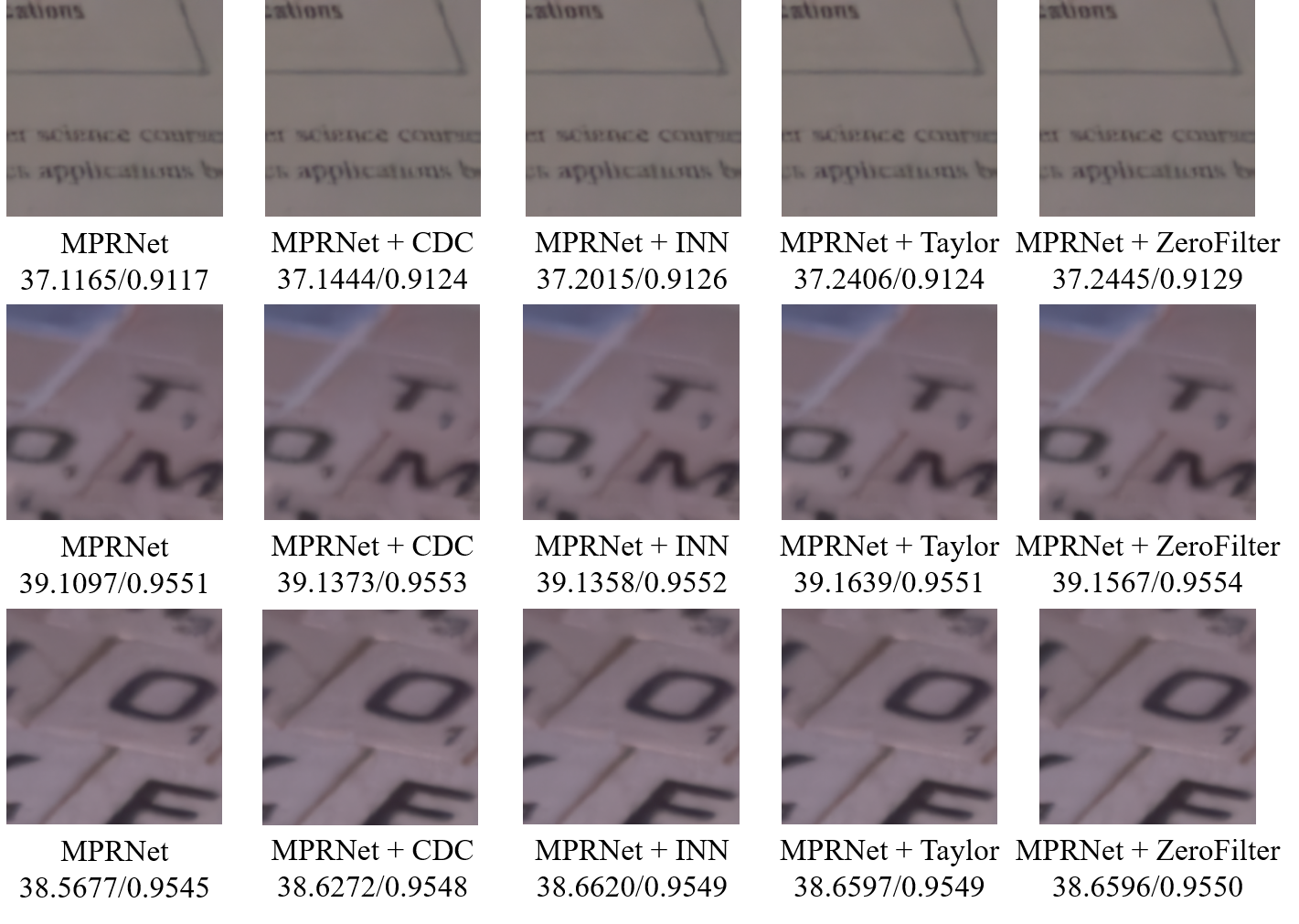}
		\caption{The visual comparison for the image de-noising. We also list the PSNR/SSIM scores under each case.}
		\label{dnmpr}
	\end{figure}

\begin{figure}[t]
		\centering
		\includegraphics[width=\columnwidth]{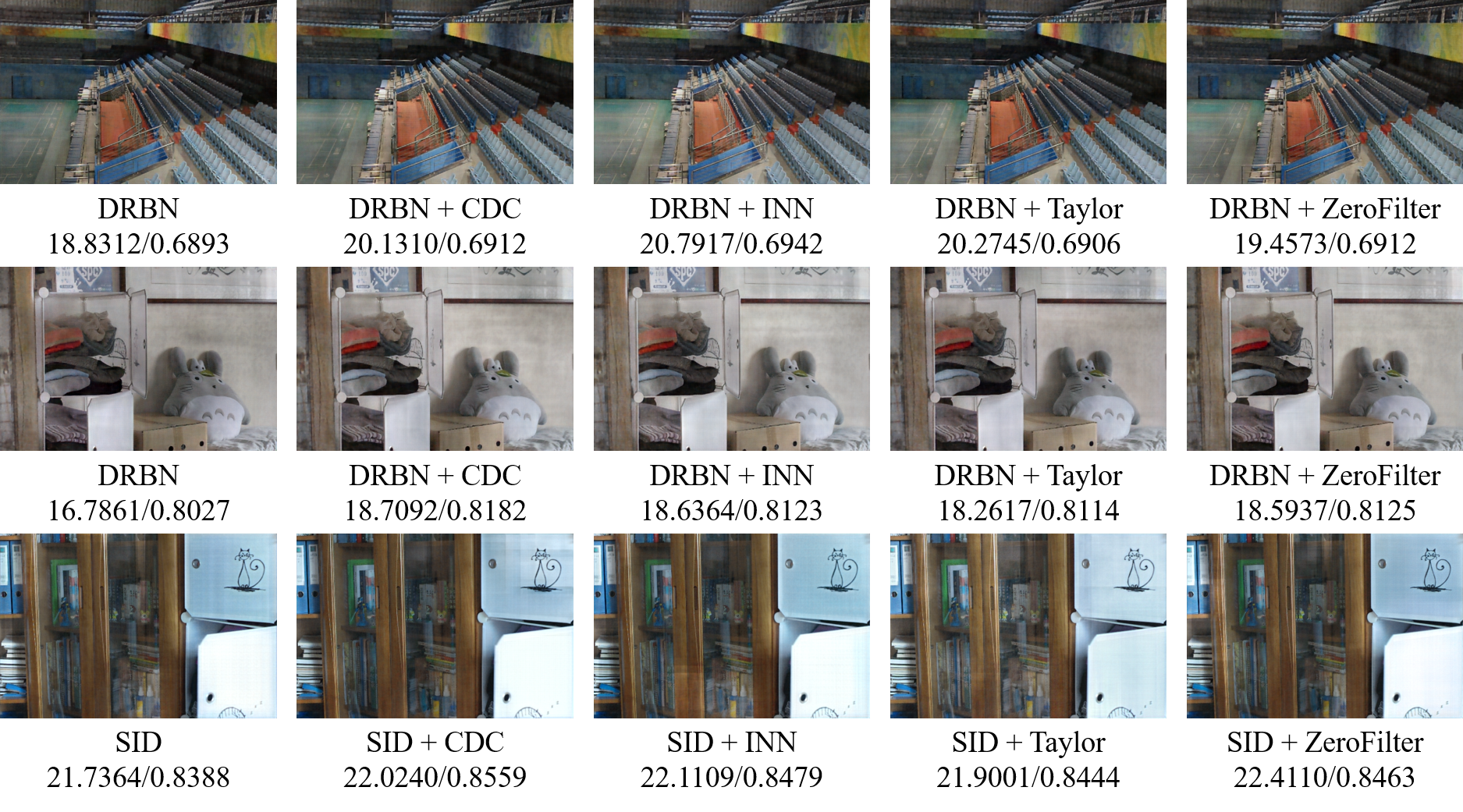}
		\caption{The visual comparison for the image enhancement. We also list the PSNR/SSIM scores under each case.}
		\label{ie}
	\end{figure}

	Due to the page limits, the main manuscript has not presented the sufficient visual results of the reported tasks over the reported baselines. In this section, we provide the representative samples to validate the effectiveness of our belief over image de-noising task of Figure \ref{dndn}, Figure \ref{dnmpr}, low-light image enhancement of Figure \ref{ie}. As can be seen, integrating with our belief is capable of improving the visual quality. 
	
	\subsection{Implementation details of ablation studies.}
	
	\textbf{Initialization strategy.} In our work, the default initialization strategy is Kaiming initialization. To explore the impact of initial mode,  we replace the default Kaiming initialization by Xavier initialization, reported in Table \ref{image de-noising as init} and Table \ref{image enhancement as init} show that replacing the default almost has little impact on performance, thus verifying the robustness of our belief.
	
	In our experiment, we select two representative random weights network manifolds by \textbf{\textit{Central Difference Convolution Manifold}} and \textbf{\textit{Invertible Neural Network Manifold}} for performance verification. In detail, we employ the Xavier initialization to weight the convolution kernels within the above manifolds.
	
	\textbf{Model architecture.} All of the loss networks are implemented by convolution network as default. To explore the architecture impact,  we replace the default CNN by Transformer. The results in Table \ref{image de-noising as arc} and Table \ref{image enhancement as arc} demonstrate that replacing it rarely affects the performance.
	
	In our experiment, we select the following random weights network manifolds by \textbf{\textit{Taylor's Unfolding Manifold}} and \textbf{\textit{Invertible Neural Network Manifold}} for performance verification. In detail, we replace the convolution part of main body part within Taylor's Unfolding Manifold by the transformer and the translation functions $F$ and $G$ within Invertible Neural Network Manifold by transformer.
	
	The reason is that 1) Reverse Filtering Network Manifolds have to stand on the low-pass filters for convergence maintaining where Multi-scale Gaussian Convolution Module is devised in our paper. Therefore, the architecture cannot change; 2) Central Difference Convolution Manifold is inborn with convolution architectures and thus cannot change. To this end, we select the above two samples.  
	
	\textbf{Model depth.}  For model depth, we change the model depth of loss network by adding the layers. To ensure a fair comparison, the other factor keeps the same. The  results in Table \ref{image de-noising as depth} and Table \ref{image enhancement as depth} demonstrate the stable performance.
	
	In our experiment, we select two representative random weights network manifolds by \textbf{\textit{Central Difference Convolution Manifold}} and \textbf{\textit{Invertible Neural Network Manifold}} for performance verification. In detail, we change the default three-layer Central Difference Convolution and Invertible Neural Network by seven layers. 
	
	\textbf{Model numbers.} In our experiment, we use the single loss network as default. As shown in Table \ref{image de-noising as num} and Table \ref{image enhancement as num}, we employ multiple parallel loss networks  to verify the impact of model numbers. The results indicates that increasing the number of models will improve the performance. It attributes to the advantages of model ensemble.
	
	In our experiment, we select two representative random weights network manifolds by \textbf{\textit{Central Difference Convolution Manifold}} and \textbf{\textit{Invertible Neural Network Manifold}} for performance verification. In detail, we change the default single loss network with three ones by 3-3-3 variants and 3-5-7 variants.
	
	\section{Limitations} 
	First, the more comprehensive experiments on broader computer vision  tasks (\emph{e.g.}, image de-blurring) have not been explored. Second, more experiments on representative baselines are missed and need to be conducted.  Note that, orthogonal to the existing data and model studies, the focus of this work is beyond proposing a novel loss function paradigm to empower the model learning capability, sparking the realms of loss function.
	
	\section{Broader impacts.}
	
	Our work shows the promising capability of the learned loss function for computer vision algorithms to empower the performance gains. Integrating our belief of learned loss function will improve the performance of neural networks and facilitate  the development of AI in real-world applications. However, the efficacy of our method may raise potential concerns when it is improperly used. For example, the safety of the applications of our  method in real-world applications may not be guaranteed. We will investigate the robustness and  effectiveness of our method in broader real-world applications.
	
	\begin{table*}[!t]
		\centering
		\renewcommand{\arraystretch}{1.221}
		\caption{\textbf{Quantitative comparisons of guided image super-resolution.}}
		\resizebox{\linewidth}{!}{ 
			\begin{tabular}{l l |c c c c  c c c c}
				\hline
				\multirow{2}{*}{Model} & \multirow{2}{*}{Configurations}& \multicolumn{4}{c}{WorldView-II} & \multicolumn{4}{c}{GaoFen2} \\
				& &PSNR$\uparrow$ & SSIM$\uparrow$&SAM$\downarrow$&ERGAS$\downarrow$ &PSNR$\uparrow$ & SSIM$\uparrow$&SAM$\downarrow$  &EGAS$\downarrow$ \\
				\hline
				\multirow{3}{*}{INNformer} & Original& 41.6903 & 0.9704 & 0.0227 & 0.9514         & 47.3528 & 0.9893 & 0.0102 & 0.5479  \\ \hline
				
				&  +Taylor & 41.8168 & 0.9716 & 0.0224 & 0.9276  & 47.4058 & 0.9901 & 0.0101 & 0.5356   \\
				
				&+CDC & 41.8072 & 0.9715 & 0.0224 & 0.9276  & 47.4121 & 0.9902 & 0.0100 & 0.5354  \\
				
				&+INN & 41.8229 & 0.9717 & 0.0223 & 0.9276  & 47.4233 & 0.9904 & 0.0100 & 0.5353  \\
				
				&+Reverse & 41.7293 & 0.9711 & 0.0226 & 0.9276  & 47.4010 & 0.9901 & 0.0101 & 0.5354  \\
				
				\hline
				
				\multirow{3}{*}{SFINet} &  Original& 41.7244 & 0.9725 & 0.0220 & 0.9506       & 47.4712 & 0.9901 & 0.0102 & 0.5462 \\ \hline
				
				& +Taylor & 41.9314 & 0.9723 & 0.0219 & 0.9278    & 47.6132 & 0.9911 & 0.0101 & 0.5277  \\
				
				& +CDC & 41.8943 & 0.9719 & 0.0220 & 0.9283     & 47.5990 & 0.9910 & 0.0101 & 0.5281  \\
				
				& +INN & 41.9521 & 0.9727 & 0.0217 & 0.9278     & 47.6316 & 0.9916 & 0.0101 & 0.5275  \\
				
				& +Reverse & 41.9217 & 0.9722 & 0.0218 & 0.9281     & 47.6227 & 0.9914 & 0.0101 & 0.5275  \\
				\hline
		\end{tabular}}
		\label{tab:ps1}
	\end{table*}

	\begin{table*}[!htb]
		\centering
		\renewcommand{\arraystretch}{1.2}
		\caption{\textbf{Ablation studies of model architecture for image enhancement.}}
		\begin{tabular}{l l |c c c c c c }
			\hline
			\multirow{2}{*}{Model} & \multirow{2}{*}{Configurations}& \multicolumn{3}{c}{LoL}   \\
			&&PSNR & SSIM & NIQE  \\
			\hline
			\multirow{5}{*}{SID} & Original& 20.2461 & 0.7920 & 4.1586   \\ \hline

			&+Taylor+epochR & 20.6018 & 0.7975 & 3.8079  \\ 
			&+Taylor+epochR+Transformer & 20.5864 & 0.7971 & 3.8348  \\

			&+INN+epochR  & 20.3958 & 0.7924 & 3.9210   \\ 
			&+INN+epochR+Transformer  & 20.3178 & 0.7944 & 3.8889   \\

			\hline

			\multirow{5}{*}{DRBN} & Original& 19.8509  & 0.7769 & 4.7738  \\ \hline

			&+Taylor+epochR & 20.2405 & 0.7791 & 4.6721 \\
			&+Taylor+epochR+Transformer & 20.1826 & 0.7784 & 4.6968 \\
			
			&+INN+epochR  & 20.1913 & 0.7769 & 4.8067 \\ 
			&+INN+epochR+Transformer  & 20.1196 & 0.7772 & 4.7163 \\
			
			\hline
		\end{tabular}
		\label{image enhancement as arc}
	\end{table*}

	\begin{table*}[t]
		\centering
		\renewcommand{\arraystretch}{1.2}
		\caption{\textbf{Ablation studies of model architecture for image de-noising.}}
		\begin{tabular}{l l |c c  c c  }
			\hline
			\multirow{2}{*}{Model} & \multirow{2}{*}{Configurations}& \multicolumn{2}{c}{SIDD}  \\
			&&PSNR$\uparrow$ & SSIM$\uparrow$     \\
			\hline
			\multirow{5}{*}{DnCNN} & Original& 37.1992 & 0.8954     \\ \hline

			&+Taylor+epochR & 37.3719  & 0.8954  \\ 
			&+Taylor+epochR+Transformer & 37.3560 & 0.8958   \\ 
			
			&+INN+epochR  & 37.3318 & 0.8964  \\  
			&+INN+epochR+Transformer  & 37.3297 & 0.8961   \\
			
			\hline
			\multirow{5}{*}{MPRnet} & Original& 39.2372
			& 0.9159   \\ \hline
			
			&+Taylor+epochR  & 39.3283  & 0.9161  \\ 
			&+Taylor+epochR+Transformer & 39.2783 & 0.9160   \\ 
			
			&+INN+epochR  & 39.3317 & 0.9162  \\  
			&+INN+epochR+Transformer  & 39.2756 & 0.9159   \\

			\hline
		\end{tabular}
		\label{image de-noising as arc}
	\end{table*}
	
	\begin{table*}[!htb]
		\centering
		\renewcommand{\arraystretch}{1.2}
		\caption{\textbf{Ablation studies of model depth for image enhancement.}}
		\begin{tabular}{l l |c c c c c c }
			\hline
			\multirow{2}{*}{Model} & \multirow{2}{*}{Configurations}& \multicolumn{3}{c}{LoL}   \\
			&&PSNR & SSIM & NIQE  \\
			\hline
			\multirow{5}{*}{SID} & Original& 20.2461 & 0.7920 & 4.1586   \\ \hline
			
			&+CDC+epochR & 20.4750 & 0.7999 & 3.6636  \\
			&+CDC(3)+epochR+Depth & 20.3464& 0.7915 & 3.8620 \\ 
			&+CDC(7)+epochR+Depth & 20.4258 & 0.7857 & 4.4067 \\

			&+INN+epochR  & 20.3858 & 0.7924 & 3.9210   \\
			&+INN(3)+epochR+Depth  & 20.4946 & 0.7862 & 4.1512   \\ 
			&+INN(7)+epochR+Depth  & 20.2816 & 0.7959 & 3.7419   \\ 
			
			\hline

			\multirow{5}{*}{DRBN} & Original& 19.8509  & 0.7769 & 4.7738  \\ \hline

			&+CDC+epochR & 20.0756 & 0.7837 & 4.7850  \\
			&+CDC(3)+epochR+Depth & 19.9188 & 0.7808 & 4.7074  \\
			&+CDC(7)+epochR+Depth & 19.9769 & 0.7795 & 4.8156  \\

			&+INN+epochR  & 20.1913 & 0.7769 & 4.8067  \\
			&+INN(3)+epochR+Depth  & 20.0330 & 0.7758 & 4.5883   \\ 
			&+INN(7)+epochR+Depth  & 20.1153 & 0.7787 & 4.7089  \\

			\hline
		\end{tabular}
		\label{image enhancement as depth}
	\end{table*}

	\begin{table*}[t]
		\centering
		\renewcommand{\arraystretch}{1.2}
		\caption{\textbf{Ablation studies of model depth for image de-noising.}}
		\begin{tabular}{l l |c c  c c  }
			\hline
			\multirow{2}{*}{Model} & \multirow{2}{*}{Configurations} & \multicolumn{2}{c}{SIDD}  \\
			&&PSNR$\uparrow$ & SSIM$\uparrow$   \\
			\hline
			\multirow{5}{*}{DnCNN} & Original& 37.1992 & 0.8954     \\ \hline

			&+CDC+epochR & 37.2784 & 0.8955   \\
			&+CDC(3)+epochR+Depth & 37.2218  & 0.8921  \\ 
			&+CDC(7)+epochR+Depth & 37.2923  & 0.8930  \\ 
			
			&+INN+epochR  & 37.3218 & 0.8964  \\
			&+INN(3)+epochR+Depth  & 37.3213 & 0.8967   \\ 
			&+INN(7)+epochR+Depth  & 37.3142 & 0.8967   \\ 
			
			\hline
			\multirow{5}{*}{MPRnet} & Original& 39.2372
			& 0.9159   \\ \hline
			
			&+CDC+epochR & 39.2821 & 0.9161  \\
			&+CDC(3)+epochR+Depth & 39.2814  & 0.9160  \\ 
			&+CDC(7)+epochR+Depth & 39.2740  & 0.9161  \\

			&+INN+epochR  & 39.2729 & 0.9162   \\
			&+INN(3)+epochR+Depth  & 39.2758 & 0.9160   \\ 
			&+INN(7)+epochR+Depth  & 39.2737 & 0.9160   \\

			\hline
		\end{tabular}
		\label{image de-noising as depth}
	\end{table*}

	\begin{table*}[!htb]
		\centering
		\renewcommand{\arraystretch}{1.2}
		\caption{\textbf{Ablation studies of model numbers for image enhancement.}}
		\begin{tabular}{l l |c c c c c c }
			\hline
			\multirow{2}{*}{Model} & \multirow{2}{*}{Configurations}& \multicolumn{3}{c}{LoL}   \\
			&&PSNR & SSIM & NIQE  \\
			\hline
			\multirow{5}{*}{SID} & Original& 20.2461 & 0.7920 & 4.1586   \\ \hline
			
			&+CDC+epochR & 20.4750 & 0.7999 & 3.6636  \\
			&+CDC+epochR+Number(357) & 20.4879 & 0.7991 & 3.6793  \\ 
			&+CDC+epochR+Number(555) & 20.5424 & 0.7889 & 3.7738  \\

			&+INN+epochR  & 20.3858 & 0.7924 & 3.9210   \\
			&+INN+epochR+Number(357)  & 20.3516 & 0.7843& 4.2365   \\  
			&+INN+epochR+Number(555)  & 20.3316 & 0.7911 & 4.1289  \\  
			
			\hline

			\multirow{5}{*}{DRBN} & Original& 19.8509  & 0.7769 & 4.7738  \\ \hline

			&+CDC+epochR & 20.0756 & 0.7837 & 4.7850  \\
			&+CDC+epochR+Number(357) & 20.0200 & 0.7789 & 4.6900  \\ 
			&+CDC+epochR+Number(555) & 20.0403 & 0.7750 & 4.7060  \\

			&+INN+epochR  & 20.1913 & 0.7769 & 4.8067   \\
			&+INN+epochR+Number(357)  & 20.0510 & 0.7779 & 4.6957   \\
			&+INN+epochR+Number(555)  & 20.2572 & 0.7767 & 4.6169   \\

			\hline
		\end{tabular}
		\label{image enhancement as num}
	\end{table*}

	\begin{table*}[t]
		\centering
		\renewcommand{\arraystretch}{1.2}
		\caption{\textbf{Ablation studies of model numbers for image de-noising.}}
		\begin{tabular}{l l |c c  c c  }
			\hline
			\multirow{2}{*}{Model} & \multirow{2}{*}{Configurations} & \multicolumn{2}{c}{SIDD}  \\
			&&PSNR$\uparrow$ & SSIM$\uparrow$     \\
			\hline
			\multirow{5}{*}{DnCNN} & Original& 37.1992 & 0.8954     \\ \hline
			
			&+CDC+epochR & 37.2784 & 0.8925   \\
			&+CDC+epochR+Number(357) & 37.4377  & 0.8969  \\ 
			&+CDC+epochR+Number(555) & 37.3208  & 0.8948  \\

			&+INN+epochR  & 37.3218 & 0.8964   \\
			&+INN+epochR+Number(357)  & 37.3374 & 0.8937   \\ 
			&+INN+epochR+Number(555)  & 37.3581 & 0.8944   \\
			
			\hline
			\multirow{5}{*}{MPRnet} & Original& 39.2372
			& 0.9159   \\ \hline

			&+CDC+epochR  & 39.2821 & 0.9162   \\
			&+CDC+epochR+Number(357)  & 39.2704 & 0.9161   \\ 
			&+CDC+epochR+Number(555)  & 39.2764 & 0.9160   \\ 
			
			&+INN+epochR & 39.2729 & 0.9162  \\
			&+INN+epochR+Number(357) & 39.2767  & 0.9160  \\ 
			&+INN+epochR+Number(555) & 39.2818 & 0.9160  \\

			\hline
		\end{tabular}
		\label{image de-noising as num}
	\end{table*}
	

	\begin{table*}[!htb]
		\centering
		\renewcommand{\arraystretch}{1.2}
		\caption{\textbf{Ablation studies of initialization strategy for image enhancement.}}
		\begin{tabular}{l l |c c c c c c }
			\hline
			\multirow{2}{*}{Model} & \multirow{2}{*}{Configurations}& \multicolumn{3}{c}{LoL}   \\
			&&PSNR & SSIM & NIQE  \\
			\hline
			\multirow{5}{*}{SID} & Original& 20.2461 & 0.7920 & 4.1586   \\ \hline
			
			&+CDC+epochR & 20.4750 & 0.7999 & 3.6636  \\
			&+CDC+epochR+xavier & 20.3271 & 0.7847 & 4.1454  \\

			&+INN+epochR  & 20.3858 & 0.7924 & 3.9210  \\
			&+INN+epochR+xavier  & 20.3257 & 0.7927 & 4.1187  \\  
			
			\hline

			\multirow{5}{*}{DRBN} & Original& 19.8509  & 0.7769 & 4.7738  \\ \hline

			&+CDC+epochR & 20.0756 & 0.7837 & 4.7850 \\
			&+CDC+epochR+xavier & 20.0136 & 0.7760 & 4.7566  \\

			&+INN+epochR  & 20.1913 & 0.7769 & 4.8067   \\
			&+INN+epochR+xavier  & 20.0948 & 0.7773 & 4.6879   \\

			\hline
		\end{tabular}
		\label{image enhancement as init}
	\end{table*}

	\begin{table*}[t]
		\centering
		\renewcommand{\arraystretch}{1.2}
		\caption{\textbf{Ablation studies of initialization strategy for image de-noising.}}
		\begin{tabular}{l l |c c  c c  }
			\hline
			\multirow{2}{*}{Model} & \multirow{2}{*}{Configurations} & \multicolumn{2}{c}{SIDD}  \\
			&&PSNR$\uparrow$ & SSIM$\uparrow$     \\
			\hline
			\multirow{5}{*}{DnCNN} & Original& 37.1992 & 0.8954     \\ \hline
			
			&+CDC+epochR & 37.2784 & 0.8925   \\
			&+CDC+epochR+xavier & 37.2567  & 0.8963 \\

			&+INN+epochR  & 37.3218 & 0.8964   \\
			&+INN+epochR+xavier  & 37.2890 & 0.8957  \\ 
			
			\hline
			\multirow{5}{*}{MPRnet} & Original& 39.2372
			& 0.9159   \\ \hline
			
			&+CDC+epochR & 39.2821 & 0.9161   \\
			&+CDC+epochR+xavier & 39.2768  & 0.9160  \\

			&+INN+epochR  & 39.2729 & 0.9162   \\
			&+INN+epochR+xavier  & 39.2779 & 0.9160  \\

			\hline
		\end{tabular}
		\label{image de-noising as init}
	\end{table*}
	
\end{document}